\documentclass[11pt]{article}

\usepackage[colorlinks,citecolor=blue]{hyperref}
\usepackage{color}
\usepackage{graphicx}
\usepackage{fancyhdr}
\usepackage{subcaption}
\usepackage{amsmath,epsfig}
\usepackage{amssymb,amsfonts,amsmath}
\usepackage{bm}
\usepackage{mathrsfs,amsmath}
\usepackage[margin=2.5cm]{geometry}

\def\Real{{\mathbb{R}}}

\def\x{{{\bf x}}}
\def\X{{{\bf X}}}
\def\C{{{\bf C}}}
\def\U{{{\bf U}}}
\def\G{{{\bf G}}}
\def\K{{{\bf K}}}
\def\A{{{\bf A}}}
\def\L{{\boldsymbol{\Lambda}}}
\def\PHI{{\boldsymbol{\Phi}}}
\def\PHi{{\boldsymbol{\phi}}}

\title{Let's consider more general nonlinear approaches to study teleconnections of climate variables}
\author{D. Bueso, M. Piles and G. Camps-Valls\\
Image Processing Laboratory (IPL), Universitat de Val\`encia, Spain\\
{\tt http://isp.uv.es -- \{diego.bueso,maria.piles,gustau.camps\}@uv.es}
}
\date{}

\begin{document}

\maketitle

\section{Introduction}

The recent work \cite{xMCA} is concerned about the problem of extracting features from spatio-temporal geophysical signals. Authors introduce the complex rotated MCA (xMCA) to deal with lagged effects and non-orthogonality of the feature representation. This method essentially (1) transforms the signals to a complex plane with the Hilbert transform; (2) applies an oblique (varimax and Promax) rotation to remove the orthogonality constraint; and (3) performs the eigendecomposition in this complex space \cite{Horel84}. We argue that this method is essentially a particular case of the method called rotated complex kernel principal component analysis (ROCK-PCA) introduced in \cite{rock}, where we proposed the same approach: first transform the data to the complex plane with the Hilbert transform and then apply the varimax rotation, with the only difference that the eigendecomposition is performed in the dual (kernel) Hilbert space \cite{Gus_09,scholkopf98}. In essence, the latter allows us to generalize the xMCA solution by extracting nonlinear (curvilinear) features when nonlinear kernel functions are used. Hence, the solution of xMCA boils down to that of ROCK-PCA when the inner product is computed in the input data space instead of in the high-dimensional (possibly infinite) kernel Hilbert space where data has been mapped to \cite{Gus_09,scholkopf98}. 

In this short correspondence we show theoretical proof that xMCA is a special case of ROCK-PCA and provide quantitative evidence that more expressive and informative features can be extracted when working with kernels; results of the decomposition of global sea surface temperature (SST) fields are shown to illustrate the capabilities of ROCK-PCA to cope with nonlinear processes, unlike xMCA. 

\section{On the theoretical equivalence}

Let us consider two datasets $\X_a\in\Real^{n\times d_a}$ and $\X_b\in\Real^{n\times d_b}$ containing samples from two random variables $A$ and $B$ with the same temporal dimension $n$ and spatial dimensions $d_a$ and $d_b$, respectively. 
The MCA maximizes the cross-covariance $C_{ab}=\frac{1}{n-1}\tilde\X_a^\top \tilde\X_b$ subject to orthogonality constraints $\U_a^\top \C_{aa}\U_a=\U_b^\top \C_{bb}\U_b=1$. 
MCA can be solved as a generalized eigenvalue problem or alternatively as a singular value decomposition (SVD). Actually, as noted in \cite{xMCA}, MCA solves exactly the same problem as the standard canonical correlation analysis (CCA) \cite{harold1936relations}, which was extended to work with more than just two datasets in \cite{kettenring1971canonical}. In practice, CCA has a limitation when working with more samples (grid spatial points) than dimensions (time steps observations), $n\ll d_a$ and $n\ll d_b$, and hence typically requires an extra regularization. This is a well-known issue in standard CCA, but does not play in favour of MCA neither as it involves the same (and very high) computational cost, i.e. that of solving an SVD with a matrix size $d_a\times d_b$. This huge computational problem can be actually addressed by working in the dual (or $R$-mode) instead of the primal ($Q$-mode) where \cite{xMCA} define it. We note here that this is not only more computationally convenient  but also allows us to generalize xMCA to nonlinear cases by means of the theory of kernels \cite{scholkopf98} and to show it is a particular case of the kernel CCA (kCCA) method \cite{lai2000kernel} directly connected to ROCK-PCA \cite{rock}. 
In what follows, we show these theoretical connections, and demonstrate that kCCA not only generalizes CCA and MCA but,  with the proper term arrangement, it also reduces to solving a single eigenproblem as in kernel PCA approaches \cite{scholkopf98,rock}. 

\subsection{MCA is equivalent to CCA}

Both MCA and CCA solve the problem of finding orthogonal projections $\U_a$ and $\U_b$ of data $\X_a$ and $\X_b$, respectively, such that the cross-covariance of the projected data is maximized:
$$\max_{\U_a,\U_b} \{(\X_a\U_a)^\top(\X_b\U_b)\}~~~~s.t.~~~(\X_a\U_a)^\top \X_a\U_a=(\X_b\U_b)^\top \X_b\U_b=1, $$
that is
$$\max_{\U_a,\U_b} \{\U_a^\top \C_{ab}\U_b\}~~~~s.t.~~~\U_a^\top \C_{aa}\U_a= \U_b^\top\C_{bb}\U_b=1.$$
One can solve this problem by singular value decomposition (SVD) of the cross-covariance matrix as suggested in \cite{xMCA}, $\C_{ab} = \U_a\boldsymbol{\Sigma}\U_b^\top$. Alternatively, one can proceed in the standard way by introducing the orthogonality constraints and then deriving and equating to zero, which leads us to a generalized eigenproblem to solve $\C_{ab}\C_{bb}^{-1}\C_{ab}^\top \U_a = \L^2\C_{aa}\U_a$, which can be better conditioned solving the eigenproblem  $\C_{aa}^{-1}\C_{ab}\C_{bb}^{-1}\C_{ab}^\top \U_a = \L^2\U_a$.
%
The equivalence between SVD and CCA has been widely studied too, and has led to interesting and controversial discussions about the appropriateness, superiority and interpretability of the feature projections computed by one method over the other, which was pointed out in \cite{newman1995caveat} and later nicely studied and summarized in \cite{cherry1996singular}. 

\subsection{Dual CCA helps in the computational cost}

Solving the SVD or the generalized eigenproblem involved in MCA is very challenging because of the needed regularization and computational cost involved. Note that in standard problems in geosciences one has more samples (grid spatial points) than dimensions (time steps observations) and hence the cross-covariance is very large, $\C_{ab}\in\Real^{d_a\times d_b}$. However, this can be addressed by operating in $R$-mode instead, as it is typically done in PCA/EOF. Let us define the projection matrices $\U_a$ and $\U_b$ in the span of the data matrices $\X_a$ and $\X_b$, that is $\U_a=\X_a^\top\U$ and $\U_b=\X_b^\top\U$, where $\U\in\Real^{n\times n}$ is the new projection matrix, with typically $n\ll d_a$ and $n\ll d_b$. 
Now, the problem 
can be rewritten as
$$\max_{\U} \{\U^\top\X_a\X_a^\top \X_b\X_b^\top\U\}~~~~s.t.~~~\U^\top\X_a\X_a^\top\X_a\X_a^\top\U=\U^\top\X_b\X_b^\top\X_b\X_b^\top\U=1,$$
that is
$$\max_{\U} \{\U_a^\top \G_{aa}\G_{bb}\U\}~~~~s.t.~~~\U^\top \G_{aa}\G_{aa}\U= \U^\top\G_{bb}\G_{bb}\U_b=1,$$
where $\G_{aa}=\X_a\X_a^\top$ and $\G_{bb}=\X_b\X_b^\top$ are the Gram matrix (inner products between the data matrices). 
One could use the SVD to solve this dual ($R$-mode) problem by simply $\G_{aa}\G_{bb} = \U_a\Sigma\U_b^\top$, or by the equivalent dual CCA problem with (properly regularized) Gram matrices as before: $\G_{aa}\G_{bb}(\G_{bb}\G_{bb})^{-1}\G_{bb}\G_{aa}\U = \L\G_{aa}\G_{aa}\U$ or  $(\G_{aa}\G_{aa})^{-1}\G_{aa}\G_{bb}(\G_{bb}\G_{bb})^{-1}\G_{bb}\G_{aa}\U = \L\U$. 
%
%
%
%
%
Obviously a great computational advantage can be achieved by solving this dual problem: only two matrices are actually computed and one (smaller) inverted. We introduced this technicality to help the use and adoption of MCA, but mainly to note that this will allow us to derive the nonlinear generalization of MCA that follows, which is equivalent to the method proposed in \cite{rock}. 

\subsection{Kernelized CCA exists and generalizes MCA}

Working with Gram matrices and this duality has been widely studied and proposed in multivariate statistics \cite{jolliffe1987rotation}. Yet, more interestingly, nonlinear versions can be derived by using similar arguments within the field of kernel methods \cite{scholkopf98,Gus_09,rojo2018digital}. The framework of kernel multivariate statistics is aimed at extracting nonlinear projections while still working with linear algebra. 
Let us first consider a feature map $\PHi: \Real^d \rightarrow \cal F$ that projects input data into a Hilbert feature space $\cal F$. The new mapped data set is defined as $\PHI = [\PHi(\x_1), \cdots, \PHi(\x_n)]^\top$, and the features extracted from the input data will now be given by $\PHI' = \PHI \U$, where matrix $\U$ is of size $\text{dim}({\cal F}) \times n_f$.

To implement practical kernel MVA algorithms we need to rewrite the equations in terms of inner products in $\cal F$ only. For doing so, we rely on the availability of two kernel matrices $\K_{aa} = \PHI_a  \PHI_a^\top$ abd $\K_{bb} = \PHI_b  \PHI_b^\top$ of dimension $n \times n$, and on the Representer's Theorem~\cite{ShaweTaylor04}, which states that the projection vectors can be written as a linear combination of the training samples, i.e, $\U = \PHI_a^\top  \A_a$, $\U = \PHI_b^\top  \A_b$, matrices $\A_a$ and $\A_b$ being the new argument for the optimization. Note the resemblance with the previous dual reformulation of the problem when working with Gram matrices instead of covariance matrices. 
This is typically referred to as the {\em kernel trick} and has been used to develop nonlinear (kernel-based) versions of standard linear methods like PCA, PLS, and CCA/MCA. The nonlinear kernel PCA (kPCA) was introduced in ~\cite{scholkopf98} and later the kernel CCA (kCCA) was introduced in~\cite{lai2000kernel}. Actually, this nonlinear generalization of CCA, and hence of MCA, is not new and has been used and studied widely \cite{mehrkanoon2017regularized,alzate2008regularized,gretton2005kernel,fukumizu2007statistical}. Many improved variants have been proposed to cope with more than two datasets in {\em multiview settings} \cite{lai2000kernel}, {\em semisupervised learning} problems \cite{blaschko2008semi,Gus_09} as well as to deal with {\em temporal domains} efficiently~\cite{biessmann2010temporal}. It is easy to show that kCCA boils down to solving  $\K_{aa}\K_{bb}(\K_{bb}\K_{bb})^{-1}\K_{bb}\K_{aa}\A = \L\K_{aa}\K_{aa}\A$ or equivalently the eigenproblem  $(\K_{aa}\K_{aa})^{-1}\K_{aa}\K_{bb}(\K_{bb}\K_{bb})^{-1}\K_{bb}\K_{aa}\A = \L\A$ with properly regularized matrices. 
%
%
%

In kernel methods one has to choose a form of the kernel function $k(\x,\x')$ that encodes the similarities between data points. It is customary to use the radial basis function (RBF) kernel, $k(\x,\x') = \exp(-\gamma\|\x-\x'\|^2)$, as it is a characteristic kernel with good theoretical and practical properties, it has only the parameter $\gamma$ to tune, and generalizes the linear kernel by capturing all higher order moments of similarity between time series. This is the kernel function we used in  \cite{rock}. 
kCCA maximizes the same objective as the linear CCA and MCA, yet in $\cal F$. The kCCA is computationally efficient in our setting and can extract nonlinear relations. Also note that when a linear (dot product) kernel is used for both domains: $\K_{aa}=\G_{aa}=\X_a\X_a^\top$ and $\K_{bb}=\G_{aa}=\X_b\X_b^\top$, one comes back to the equivalent yet more efficient CCA/MCA introduced before.  
Therefore, kCCA generalizes CCA (or MCA) and the solution is the same as in the KPCA schemes in \cite{scholkopf98,rock}. 


\subsection{Remarks}

The proposed xMCA in \cite{xMCA} resolves MCA (or alternatively CCA) in the complex plane and adds an oblique extra rotation therein, as proposed earlier in \cite{rock}. These  important operations do not change the rationale that connects xMCA and ROCK-PCA because both methods operate with the same data. The previous demonstrations can be actually derived with complex data and after a rotation in Hilbert spaces: the same arguments and equations are valid when working with complex algebra in Hilbert spaces, cf. \cite{bouboulis2010extension}, and a rotation in Hilbert spaces can be expressed solely in terms of kernel evaluations too \cite{scholkopf98,rojo2018digital}.

\section{On the information content of the components}

A classical way to evaluate dimensionality reduction methods, like PCA or CCA, is to quantify how much variance of the data is explained by retaining a number of principal components. A more sensible criterion is {\em information}, but its estimation is complicated in multivariate and high-dimensional problems  because it typically implies multivariate density estimation. Nevertheless, kernel methods have provided an adequate framework to estimate independence between random variables and hence to quantify information \cite{GreBouSmoSch05}. Interestingly, there is a tight connection between kernel-dependence estimation and kCCA; in~\cite{BacJor02}, a regularized correlation operator was derived from the covariance and cross-covariance operators, and its largest singular value (the kernel canonical correlation, or kCCA) was used as a statistic to test independence. Later, in~\cite{GreSmoBouHeretal05}, the largest singular value of the cross-covariance operator, called the constrained covariance (COCO)~\cite{GreSmoBouHeretal05}, was proposed as an efficient alternative to kCCA that needs no regularization. A variety of empirical kernel quantities derived from bounds on the mutual information that hold near independence were also proposed, namely the kernel Generalised Variance (kGV) and the Kernel Mutual Information (kMI)~\cite{Gretton05}. Later, HSIC~\cite{Gretton05COLT} was introduced as an extension of COCO that allows considering the entire spectrum of the cross-covariance operator, not just the largest singular value. An illustrative example of these measures using synthetic data is given in Table~\ref{correlationdependence}, which evidences that linear methods (like correlation, CCA and MCA) fail in estimating dependence, while their kernel versions capture the nonlinear data structure and thus provide better information measures. 
We show in Fig.~\ref{fig:kgv} how a nonlinear version of CCA/MCA (kCCA) can yield more information for a wide range of values of the kernel length scale $\sigma$, thus generalizing its linear counterpart.

\begin{table}[t!]
\begin{center}
\small
\setlength{\tabcolsep}{1pt}
\begin{tabular}{l|ccc}
&
\includegraphics[height=3.4cm]{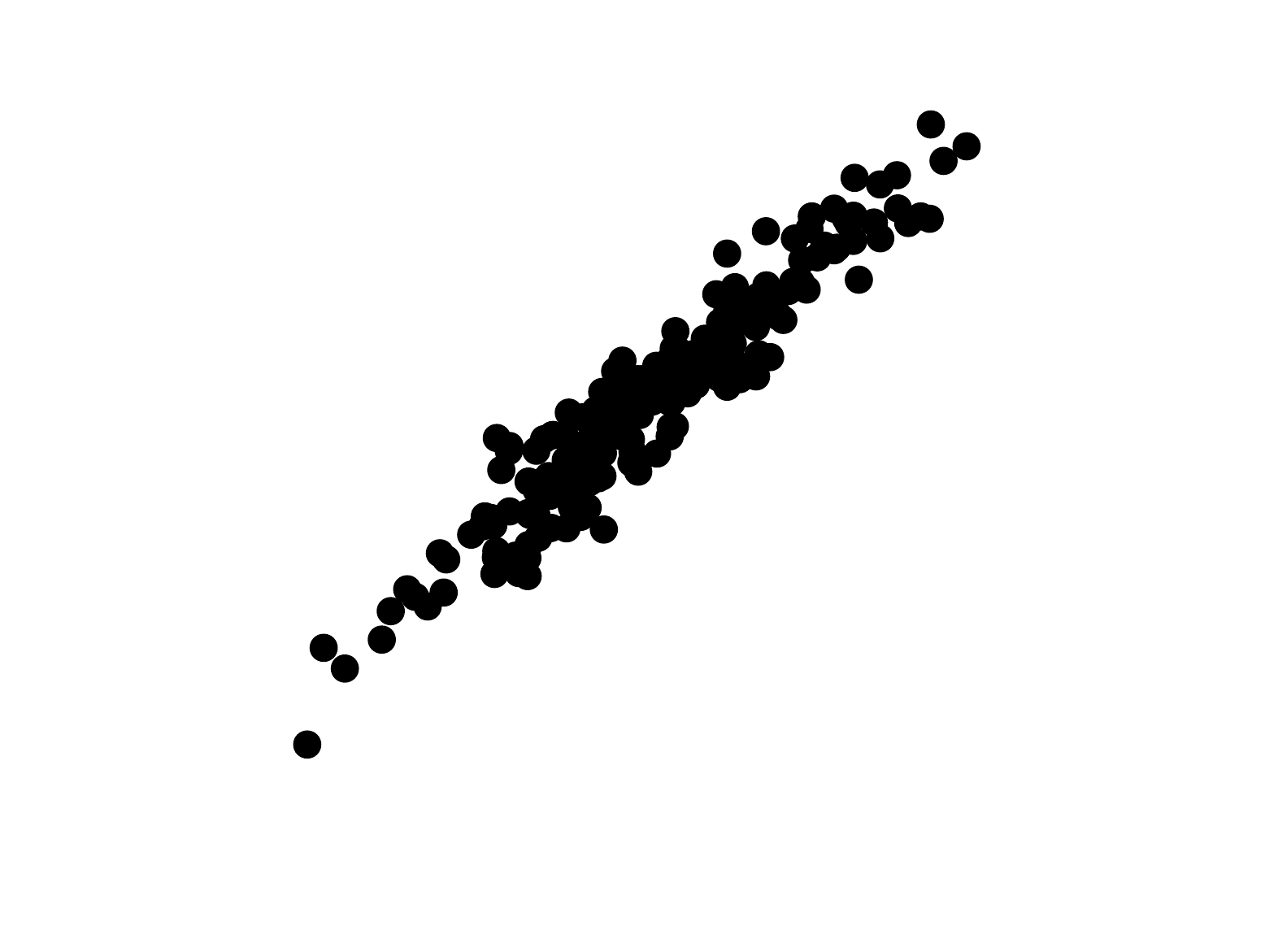}  & 
\includegraphics[height=3.4cm]{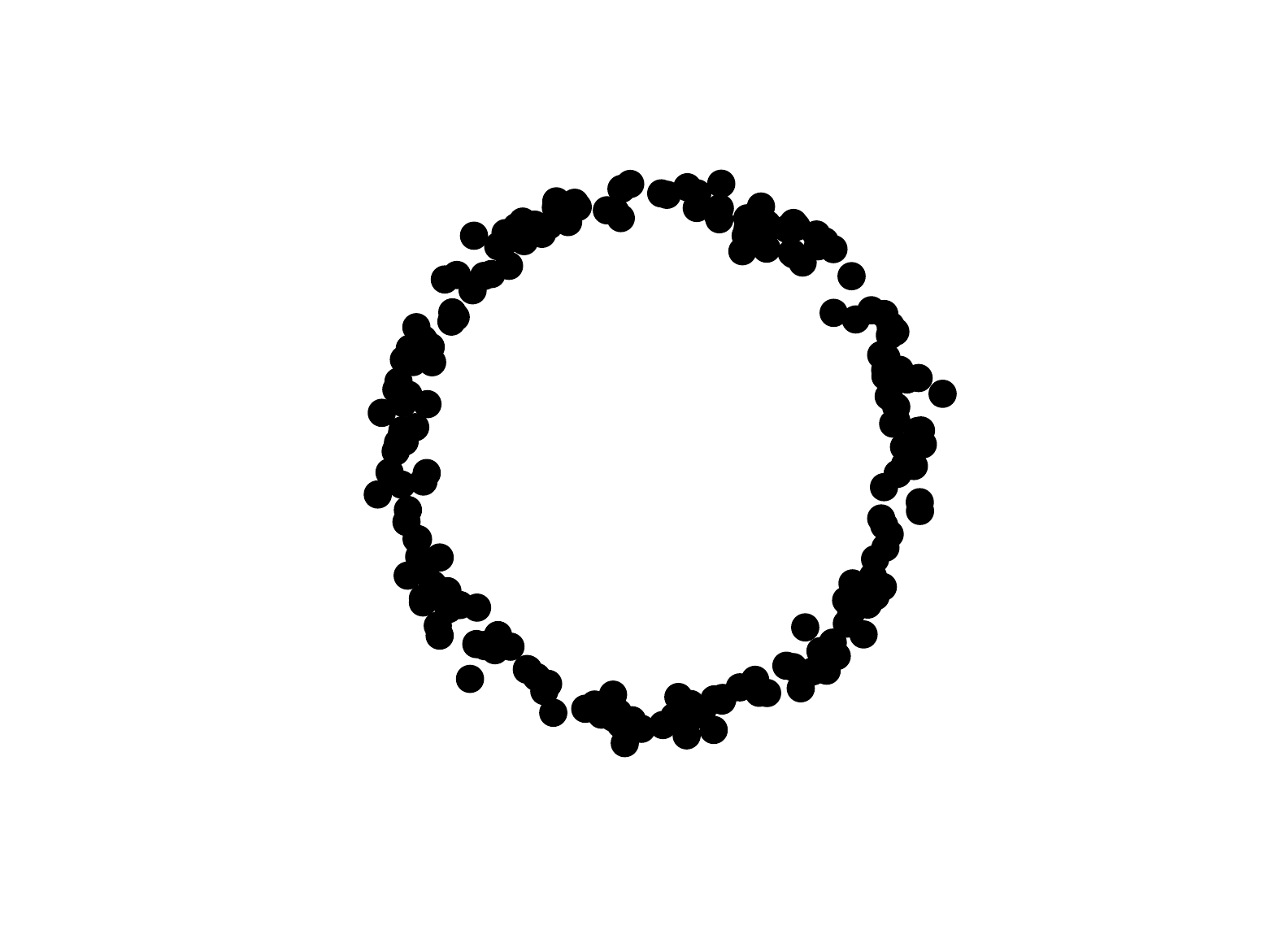}   &
\includegraphics[height=3.4cm]{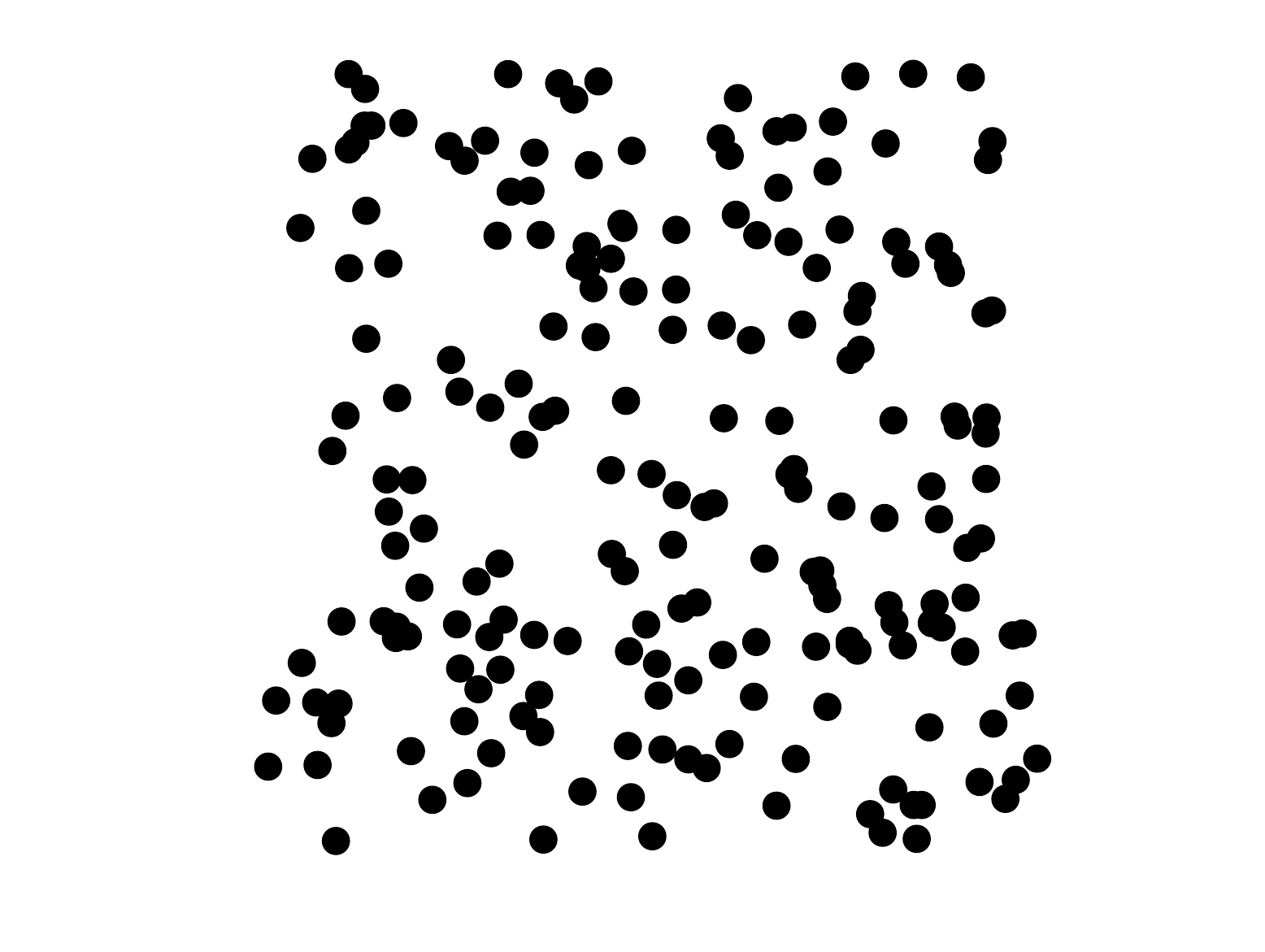} \\
\hline
Pearson's R           & 0.94  &  0.04  & 0.06  \\
MCA                   & 1.64  &  0.00  & 0.00  \\
CCA                   & 0.94  &  0.04  & 0.06   \\
\hline
HSIC (linear kernels) & 3.42  &  0.25  &  0.00  \\
HSIC (RBF kernels)    & 21.32  &  13.49  &  0.86 \\
kGV                   & 5.82  &  4.85  &  1.44  \\
\hline
kCCA & 0.95  &  0.96   &  0.42  \\
\end{tabular}
\end{center}
\caption{Dependence estimates for three examples revealing high correlation (and hence high dependence) [left], high dependence but null correlation [middle], and zero correlation and dependence [right]. The Pearson's correlation coefficient R, MCA, CCA and linear HSIC only capture second order statistics while the rest capture in general higher order dependencies.
\label{correlationdependence}}
\end{table}

\begin{figure}[ht!]
\begin{center}
\setlength{\tabcolsep}{2pt}
\begin{tabular}{ccc}
\includegraphics[width=5.3cm]{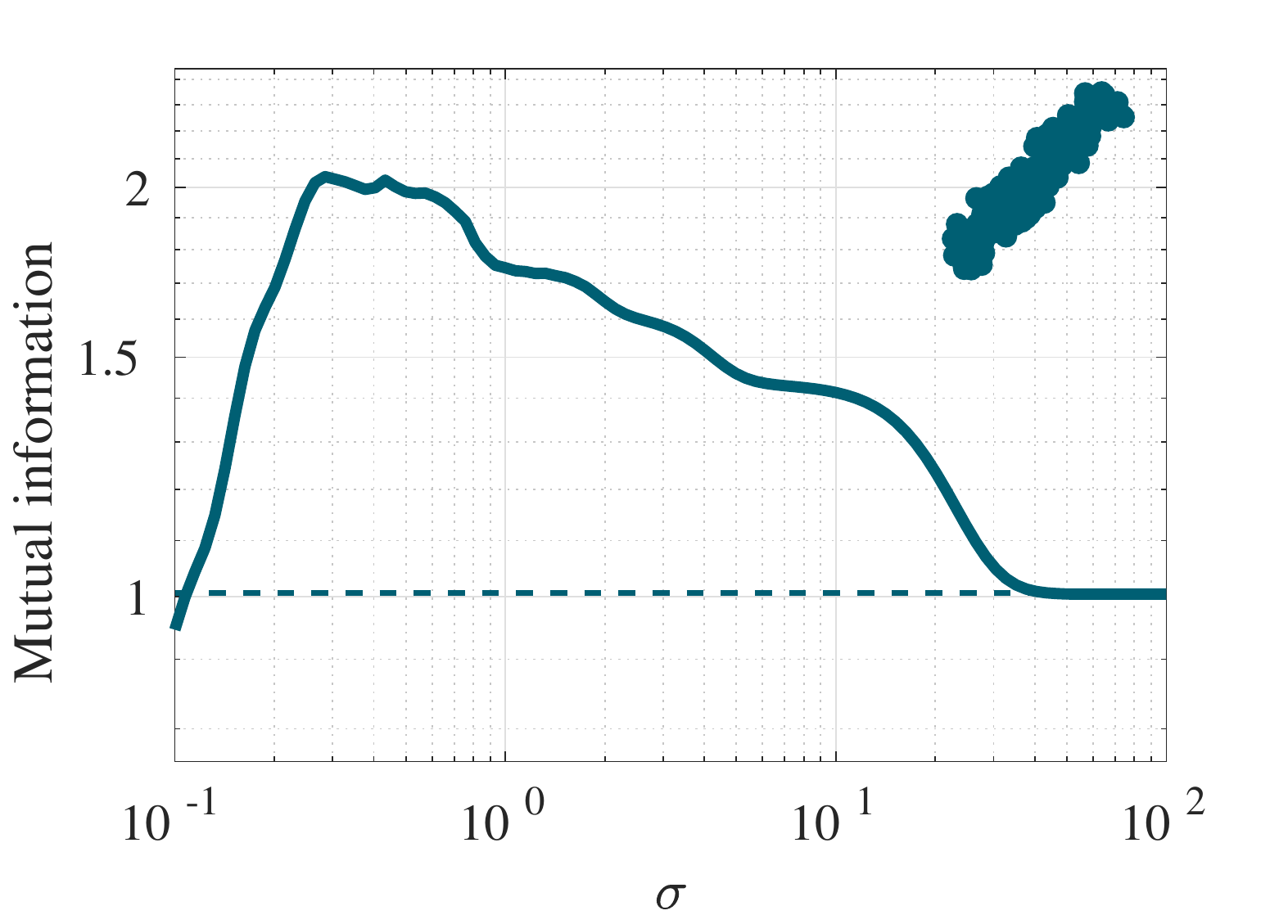}  &
\includegraphics[width=5.3cm]{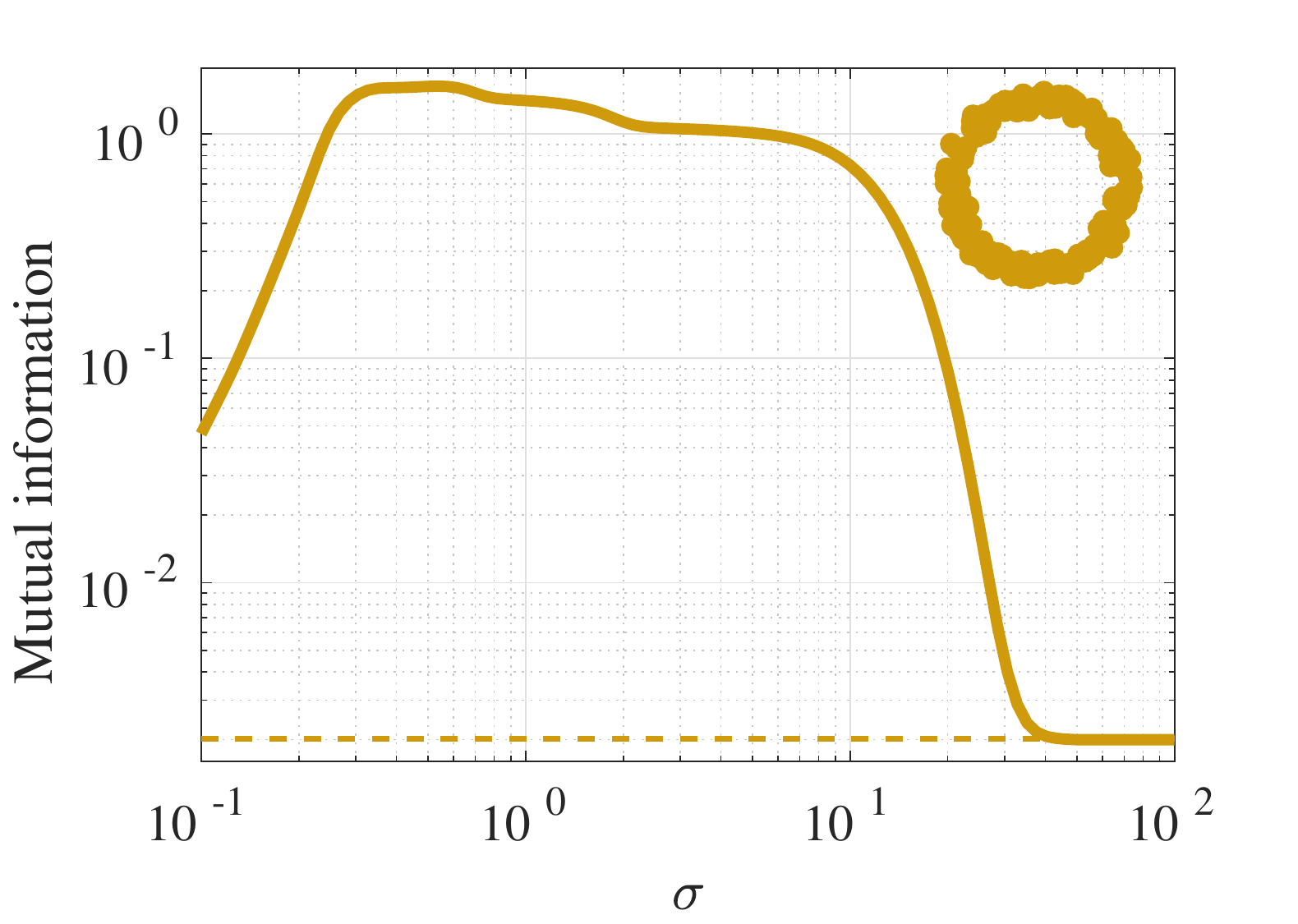} &
\includegraphics[width=5.3cm]{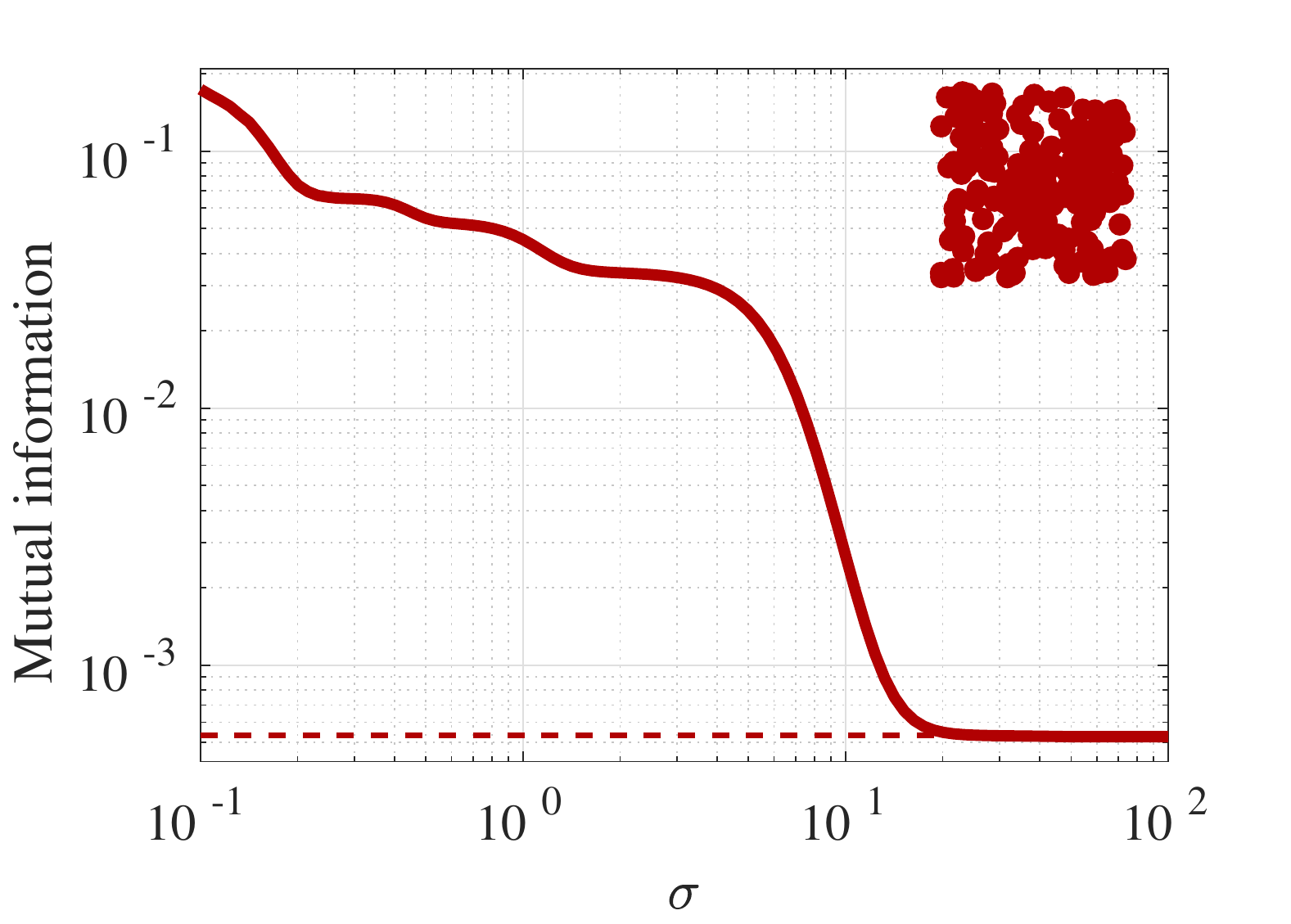}
\end{tabular}
\end{center}
\vspace{-0.5cm}
\caption{Mutual information as a function of the kernel length scale $\sigma$ for both linear ($p=1$, dashed) and kernel kCCA (nonlinear CCA/MCA). As the $\sigma$ value increases, the nonlinear version approximates the solution of the linear one in all three cases. There is an optimal $\sigma$ to capture both correlation (left plot) and dependence (middle plot)  while for the independence case (right plot) the difference is insignificant (note the low values of mutual information) through the kernel length scale $\sigma$ domain.} \label{fig:kgv}
\end{figure}

\section{An empirical comparison}

We showed that a kernel method generalizes the performance and yields more expressive (informative) features than its linear counterpart, while still relying on linear algebra. Yet, one may ask why  nonlinear components are necessary in the study of teleconnections and time-lagged correlations in the geosciences in practice. We illustrate this with an experiment decomposing the global sea surface temperature (SST) from the MetOffice reanalysis HadISST1 \cite{Parker2003}. The same example was introduced in  \cite{xMCA} with linear CMA and previously in \cite{rock} with the nonlinear PCA generalization, both operating in the complex domain and with an extra oblique rotation. 

\begin{figure*}[ht!]
\begin{center}
\setlength{\tabcolsep}{2pt}
\begin{tabular}{cc}
\includegraphics[width=16cm]{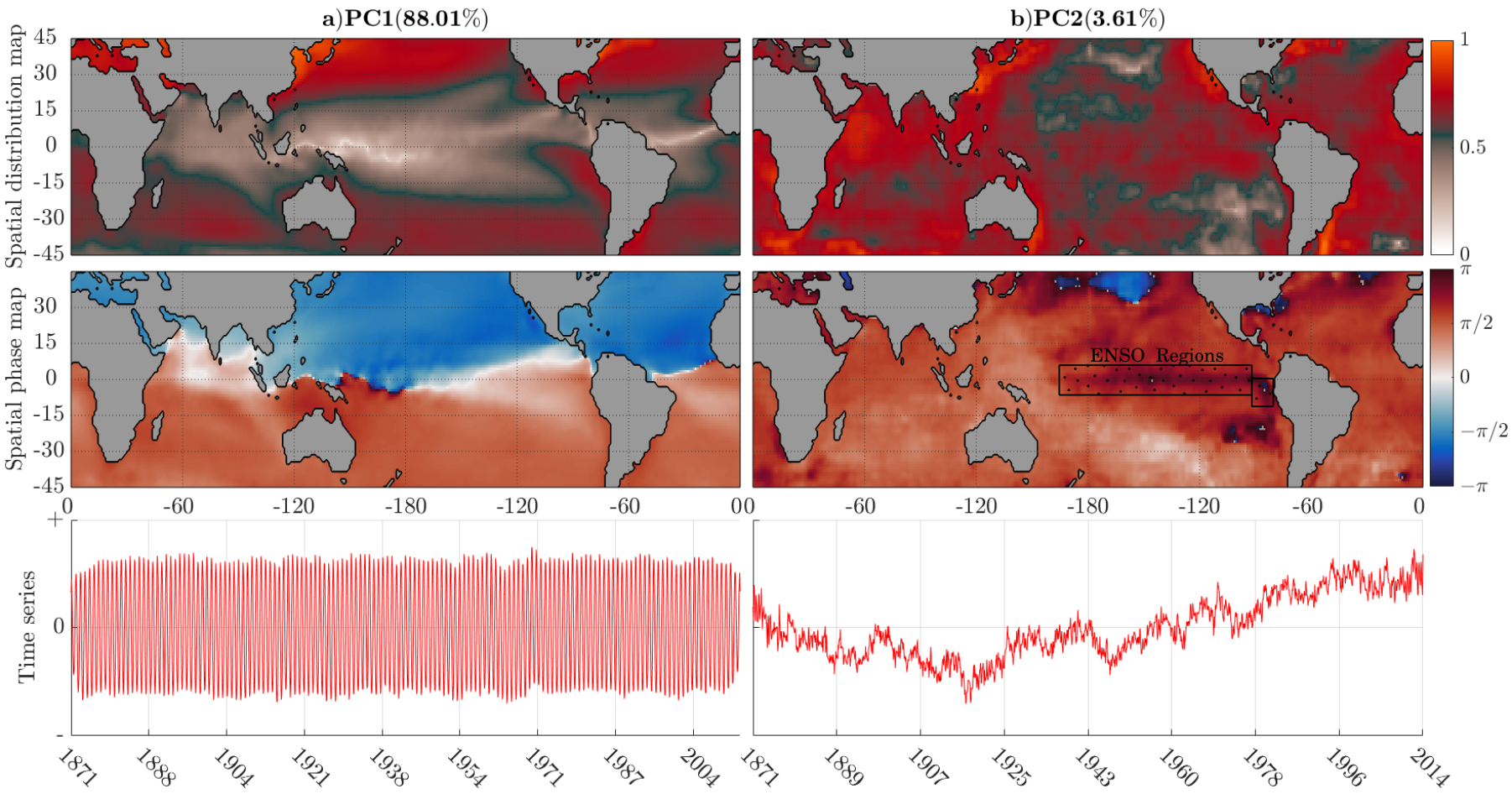}
\end{tabular}
\end{center}
\vspace{-0.5cm}
\caption{
Global SST decomposition results by ROCK-PCA (adapted from \cite{rock}), cf. compare to Fig. 5 and Fig. 8 in \cite{xMCA}. Spatial distribution of the first (a) and second (b) components for both the spatial covariation amplitude normalized to one (top row), the spatial covariation phase (middle row), and the temporal component (bottom row). ENSO regions are also shown.
} \label{fig:fig2}
\end{figure*}

\begin{figure*}[ht!]
\begin{center}
\setlength{\tabcolsep}{2pt}
\begin{tabular}{c}
\includegraphics[width=14cm]{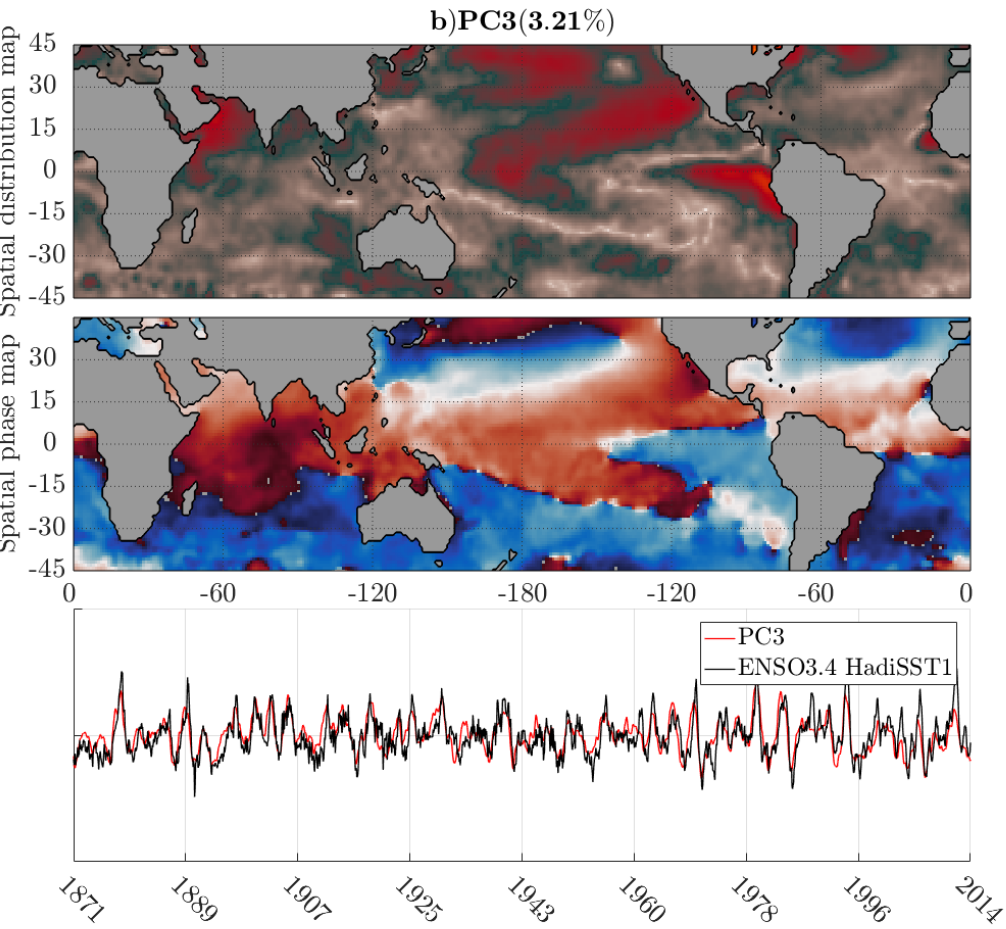}
\end{tabular}
\end{center}
\vspace{-0.5cm}
\caption{Spatial distribution of the amplitude normalized of the third component (top), the spatial covariation phase (middle) and the temporal component (bottom) by ROCK-PCA \cite{rock}. The ENSO3.4 index is highly correlated with the real component of the third component ($\rho=0.81$). Compare to Fig. 10 in \cite{xMCA} where the PC is highly correlated to the ENSO Modoki index.
} \label{fig:fig3}
\end{figure*}

Decomposition of the spatio-temporal datacube with each method allows us to extract similar principal components of SST with only very minor differences. The explained variance is more scattered across the eigenspectrum with the xMCA that with ROCK-PCA: for example, the explained variance of the first principal component (PC1) corresponding to the annual oscillation is $62.7\%$ for xMCA and $88.1\%$ for the ROCKPCA. This is an under-representation issue in  xMCA, since the annual amplitude is close to 8.5$^\circ$C and the inter-annual variability not larger than 0.11$^\circ$C per decade, with a global average increase of 0.3$^\circ$ since 1960 until 2020 \cite{Bulgin2020, chen2010}. This causes an unrealistic attribution of the relevance to the interdecadal trend (PC3 of xMCA from \cite{xMCA}), being the $6.9\%$ of the overall variance. This difference in the attribution of the explained variance emerges from the definition of the MCA, since the extracted eigenvalues do not match the individual explained variance of the separated phenomenon for more than a single variable.
The extracted components in both cases have the same physical interpretation, see figures \ref{fig:fig2} and \ref{fig:fig3} from \cite{rock}, but with difference into the mixture of components. 

Rotation of principal components with Varimax or Promax allows us to linearly maximize variance representation of a subset of principal components. This can lead to a more representative subset modes of variation if the represented physical phenomena are linearly separable. Yet, a nonlinear spatio-temporal mixture of physical phenomenon will not be separable by a linear spatio-temporal decomposition and an extra rotation as in xMCA. In this case, we can see that xMCA identifies the third PC as the interdecadal trend, but the spatial distribution of the variability is mixed with other patterns as the ENSO spatial pattern, which should exhibit a more homogeneous distribution \cite{Bulgin2020}. Figure \ref{fig:fig2} shows in the right panel the identified interdecadal trend by ROCK-PCA instead, showing a more consistent spatial pattern as previously reported elsewhere \cite{Bulgin2020}.

\begin{figure*}[ht!]
\begin{center}
\setlength{\tabcolsep}{2pt}
\begin{tabular}{cc}
\includegraphics[width=7.5cm]{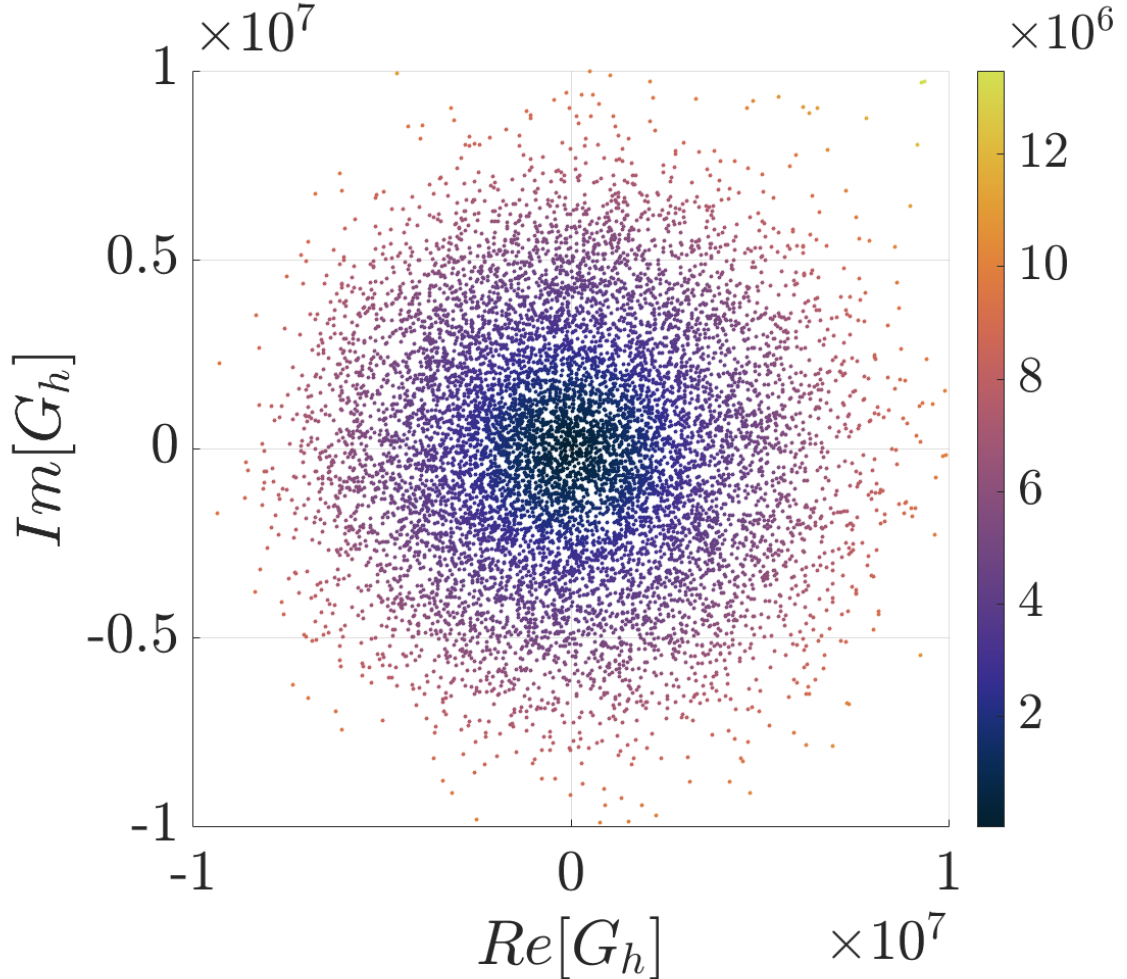} & [2pt]
\includegraphics[width=7.5cm]{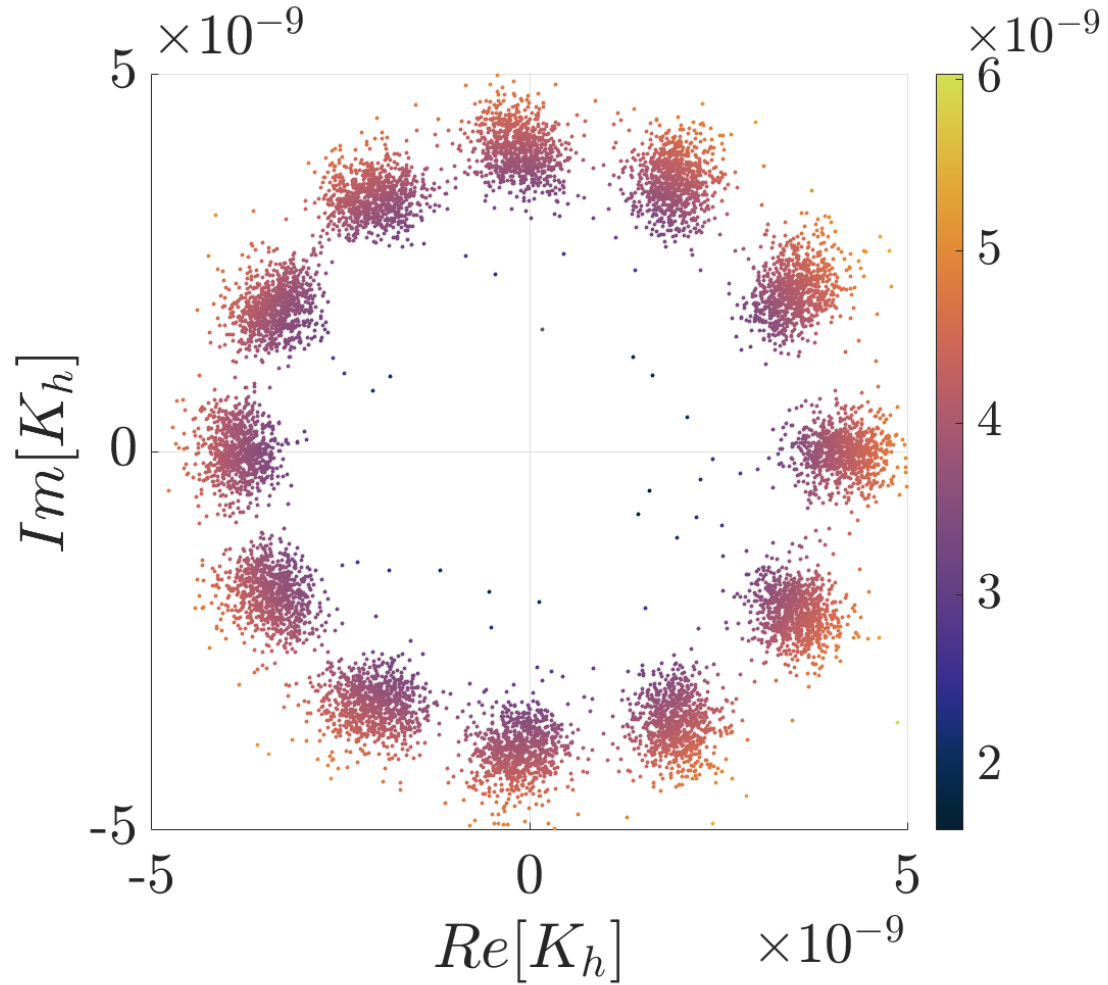}
\end{tabular}
\end{center}
\vspace{-0.5cm}
\caption{
Scatter plot of real and imaginary components of linear (left) and nonlinear (right) covariance matrix of SST.
Mapping data into a kernel Hilbert space allows us to  extract the relation of monthly variance in semantic and compact clusters. 
} \label{fig:fig4}
\end{figure*}

 In Figure \ref{fig:fig4}, we show the difference between the linear and the nonlinear case of the estimated covariance matrix, where the scatter plots show how the complex components differentiate the data into monthly variability obtained with ROCK-PCA (right) unlike in the linear case (left). Adding the nonlinear mapping of the data, we can differentiate the different physical phenomena, which a linear rotation cannot disentangle.
These results support the use of more general nonlinear approaches to decompose climatic data since they allow unraveling physical phenomena mixed in spatio-temporal data and thus extract more meaningful and informative representations. 

\section{Outlook}

Both ROCK-PCA \cite{rock} and xMCA \cite{xMCA} extract features in a complex domain after an additional oblique rotation is applied, which were proposed to deal with time-lagged processes of spatio-temporal data. Distinctly, ROCK-PCA can extract nonlinear features as it operates in a reproducing kernel Hilbert space. 
We showed that xMCA is a particular case of kernel CCA when assuming linearity. We also showed the tight connection between xMCA and ROCK-PCA. Working in the dual domain ($R$-mode) was not only computationally advantageous but also allowed us to derive the nonlinear version of xMCA which, after reformulation as a eigenproblem, reduces to solving ROCK-PCA when one dataset is involved. ROCK-PCA is not only more efficient computationally, but also provides more expressive and informative components, which permit capturing meaningful physical processes more accurately. 
Using a linear kernel is indeed quite restrictive in general, as it has been demonstrated widely that components and modes of variability in Earth and climatic data show nonlinear relations. Actually, in other works, we explored the potential use of ROCK-PCA to characterize nonlinear cross-information between covariates and global teleconnections \cite{CI19} and to identify causal footprints of climate phenomena \cite{xkgc}. In all these cases, nonlinear kernels were strictly necessary to achieve physically meaningful results. We hope that this clarification will place the novelty and limitations of xMCA in context, motivate the use of nonlinear approaches to study teleconnections, and ultimately encourage contributions in this exciting field of statistical analysis of spatio-temporal Earth and climate data. 

\section*{Acknowledgments}

This work was supported by the "Understanding and Modeling of the Earth System with Machine Learning" (USMILE) under the European Research Council (ERC) Synergy Grant 2019, Agreement N$^\circ$ 855187.


\begin{thebibliography}{10}

\bibitem{xMCA}
Niclas Rieger, \'Alvaro Corral, Estrella Olmedo, and Antonio Turiel.
\newblock Lagged teleconnections of climate variables identified via complex
  rotated maximum covariance analysis.
\newblock {\em Journal of Climate}, 34(24):9861 -- 9878, 2021.

\bibitem{Horel84}
J.~D. Horel.
\newblock Complex principal component analysis: Theory and examples.
\newblock {\em Journal of Applied Meteorology and Climatology}, 23(12):1660 --
  1673, 1984.

\bibitem{rock}
Diego Bueso, Maria Piles, and Gustau Camps-Valls.
\newblock Nonlinear {PCA} for spatio-temporal analysis of earth observation
  data.
\newblock {\em IEEE Transactions on Geoscience and Remote Sensing},
  58(8):5752--5763, 2020.

\bibitem{Gus_09}
Jerónimo Arenas-Garc\'ia and Kaare Brandt~Petersen.
\newblock {\em Kernel Multivariate Analysis in Remote Sensing Feature
  Extraction}, chapter~14, pages 327--352.
\newblock John Wiley \& Sons, Ltd, 2009.

\bibitem{scholkopf98}
Bernhard Sch{\"o}lkopf, Alexander Smola, and Klaus-Robert M{\"u}ller.
\newblock {Nonlinear Component Analysis as a Kernel Eigenvalue Problem}.
\newblock {\em Neural Computation}, 10(5):1299--1319, 07 1998.

\bibitem{harold1936relations}
Hotelling Harold.
\newblock Relations between two sets of variates.
\newblock {\em Biometrika}, 28(3/4):321--377, 1936.

\bibitem{kettenring1971canonical}
Jon~R Kettenring.
\newblock Canonical analysis of several sets of variables.
\newblock {\em Biometrika}, 58(3):433--451, 1971.

\bibitem{lai2000kernel}
Pei~Ling Lai and Colin Fyfe.
\newblock Kernel and nonlinear canonical correlation analysis.
\newblock {\em International Journal of Neural Systems}, 10(05):365--377, 2000.

\bibitem{newman1995caveat}
Matthew Newman and Prashant~D Sardeshmukh.
\newblock A caveat concerning singular value decomposition.
\newblock {\em Journal of Climate}, 8(2):352--360, 1995.

\bibitem{cherry1996singular}
Steve Cherry.
\newblock Singular value decomposition analysis and canonical correlation
  analysis.
\newblock {\em Journal of Climate}, 9(9):2003--2009, 1996.

\bibitem{jolliffe1987rotation}
Ian~T Jolliffe.
\newblock Rotation of principal components: some comments.
\newblock {\em Journal of Climatology}, 7(5):507--510, 1987.

\bibitem{rojo2018digital}
Jos{\'e}~Luis Rojo-{\'A}lvarez, Manel Mart{\'\i}nez-Ram{\'o}n, Jordi
  Munoz-Mari, and Gustau Camps-Valls.
\newblock {\em Digital signal processing with Kernel methods}.
\newblock John Wiley \& Sons, 2018.

\bibitem{ShaweTaylor04}
John Shawe-Taylor and Nello Cristianini.
\newblock {\em Kernel Methods for Pattern Analysis}.
\newblock Cambridge University Press, 2004.

\bibitem{mehrkanoon2017regularized}
Siamak Mehrkanoon and Johan~AK Suykens.
\newblock Regularized semipaired kernel cca for domain adaptation.
\newblock {\em IEEE transactions on neural networks and learning systems},
  29(7):3199--3213, 2017.

\bibitem{alzate2008regularized}
Carlos Alzate and Johan~AK Suykens.
\newblock A regularized kernel cca contrast function for ica.
\newblock {\em Neural Networks}, 21(2-3):170--181, 2008.

\bibitem{gretton2005kernel}
Arthur Gretton, Ralf Herbrich, Alexander Smola, Olivier Bousquet, Bernhard
  Sch{\"o}lkopf, et~al.
\newblock Kernel methods for measuring independence.
\newblock {\em JMLR}, 2005.

\bibitem{fukumizu2007statistical}
Kenji Fukumizu, Francis~R Bach, and Arthur Gretton.
\newblock Statistical consistency of kernel canonical correlation analysis.
\newblock {\em Journal of Machine Learning Research}, 8(2), 2007.

\bibitem{blaschko2008semi}
Matthew~B Blaschko, Christoph~H Lampert, and Arthur Gretton.
\newblock Semi-supervised laplacian regularization of kernel canonical
  correlation analysis.
\newblock In {\em Joint European conference on machine learning and knowledge
  discovery in databases}, pages 133--145. Springer, 2008.

\bibitem{biessmann2010temporal}
Felix Bie{\ss}mann, Frank~C Meinecke, Arthur Gretton, Alexander Rauch, Gregor
  Rainer, Nikos~K Logothetis, and Klaus-Robert M{\"u}ller.
\newblock Temporal kernel cca and its application in multimodal neuronal data
  analysis.
\newblock {\em Machine Learning}, 79(1):5--27, 2010.

\bibitem{bouboulis2010extension}
Pantelis Bouboulis and Sergios Theodoridis.
\newblock Extension of wirtinger's calculus to reproducing kernel hilbert
  spaces and the complex kernel lms.
\newblock {\em IEEE Transactions on Signal Processing}, 59(3):964--978, 2010.

\bibitem{GreBouSmoSch05}
A.~Gretton, O.~Bousquet, A.~J. Smola, and B.~Sch\"olkopf.
\newblock Measuring statistical dependence with {H}ilbert-{S}chmidt norms.
\newblock In {\em ALT}, pages 63--77, 2005.

\bibitem{BacJor02}
Francis Bach and Michael~I. Jordan.
\newblock Kernel independent component analysis.
\newblock {\em Journal of Machine Learning Research}, 3:1--48, 2002.

\bibitem{GreSmoBouHeretal05}
A.~Gretton, A.~Smola, O.~Bousquet, R.~Herbrich, A.~Belitski, M.~Augath,
  Y.~Murayama, J.~Pauls, B.~Sch\"olkopf, and N.~Logothetis.
\newblock Kernel constrained covariance for dependence measurement.
\newblock In Robert~G. Cowell and Zoubin Ghahramani, editors, {\em Proceedings
  of the Tenth International Workshop on Artificial Intelligence and
  Statistics}, pages 112--119, New Jersey, 2005. Society for Artificial
  Intelligence and Statistics.

\bibitem{Gretton05}
A.~Gretton, R.~Herbrich, and A.~Hyv{\"a}rinen.
\newblock Kernel methods for measuring independence.
\newblock {\em Journal of Machine Learning Research}, 6:2075--2129, 2005.

\bibitem{Gretton05COLT}
A.~Gretton, O.~Bousquet, A.~Smola, and B.~Sch{\"o}lkopf.
\newblock Measuring statistical dependence with hilbert-schmidt norms.
\newblock In {\em Proc. 16th Intl. Conf. Algorithmic Learning Theory}, pages
  63--77, Springer, 2005. Springer.

\bibitem{Parker2003}
N.~A. Rayner, D.~E. Parker, E.~B. Horton, C.~K. Folland, L.~V. Alexander, D.~P.
  Rowell, E.~C. Kent, and A.~Kaplan.
\newblock Global analyses of sea surface temperature, sea ice, and night marine
  air temperature since the late nineteenth century.
\newblock {\em Journal of Geophysical Research: Atmospheres}, 108(D14), 2003.

\bibitem{Bulgin2020}
Claire~E. Bulgin, Christopher~J. Merchant, and David Ferreira.
\newblock Tendencies, variability and persistence of sea surface temperature
  anomalies.
\newblock {\em Scientific Reports}, 1(10):7986, 5 2020.

\bibitem{chen2010}
Ge~Chen, Baomin Shao, Yong Han, Jun Ma, and Bertrand Chapron.
\newblock Modality of semiannual to multidecadal oscillations in global sea
  surface temperature variability.
\newblock {\em Journal of Geophysical Research: Oceans}, 115(C3), 2010.

\bibitem{CI19}
Diego Bueso, Maria Piles, and Gustau Camps-Valls.
\newblock Cross-information kernel causality: Revisiting global teleconnections
  of enso over soil moisture and vegetation.
\newblock In {\em Proceedings of the 9th International Workshop on Climate
  Informatics: CI 2019}, Climate Informatics, pages 172--176, Paris, 2019.
  NCAR.

\bibitem{xkgc}
Diego Bueso, Maria Piles, and Gustau Camps-Valls.
\newblock Explicit granger clkopf98ausality in kernel hilbert spaces.
\newblock {\em Phys. Rev. E}, 102:062201, Dec 2020.

\end{thebibliography}

\end{document}